\documentclass{styles/svproc}

\usepackage{url}

\usepackage[misc]{ifsym}
\usepackage{amsmath}
\DeclareMathOperator{\Tr}{Tr}
\usepackage{graphicx}
\usepackage{siunitx}
\usepackage{hyperref}
\usepackage{lipsum}

\begin{document}
\mainmatter              
\title{Improved Techniques for the Conditional Generative Augmentation of Clinical Audio Data}
\titlerunning{Improved Generative Data Augmentation}  
%
\author{Mane Margaryan$^{1, *}$, Matthias Seibold$^{1,2,*,}$\textsuperscript{\Letter}, Indu Joshi$^{1}$, Mazda Farshad$^3$, Philipp Fürnstahl$^2$, Nassir Navab$^1$}
\authorrunning{Mane Margaryan, Matthias Seibold et al.} 

\institute{Computer Aided Medical Procedures, Technical University Munich, Germany\\
\and
Research in Orthopedic Computer Science, Balgrist University Hospital, University of Zurich, Switzerland \\
\and
Department of Orthopedics, Balgrist University Hospital, University of Zurich, Switzerland \\
\Letter \, \email{matthias.seibold@balgrist.ch} \\
$\ast$ equally contributing first authors in alphabetical order
}

\maketitle

\begin{abstract}

Data augmentation is a valuable tool for the design of deep learning systems to overcome data limitations and stabilize the training process. Especially in the medical domain, where the collection of large-scale data sets is challenging and expensive due to limited access to patient data, relevant environments, as well as strict regulations, community-curated large-scale public datasets, pretrained models, and advanced data augmentation methods are the main factors for developing reliable systems to improve patient care. However, for the development of medical acoustic sensing systems, an emerging field of research, the community lacks large-scale publicly available data sets and pretrained models. To address the problem of limited data, we propose a conditional generative adversarial neural network-based augmentation method which is able to synthesize mel spectrograms from a learned data distribution of a source data set. In contrast to previously proposed fully convolutional models, the proposed model implements residual Squeeze and Excitation modules in the generator architecture. We show that our method outperforms all classical audio augmentation techniques and previously published generative methods in terms of generated sample quality and a performance improvement of $2.84\%$ of Macro F1-Score for a classifier trained on the augmented data set, an enhancement of $1.14\%$ in relation to previous work. By analyzing the correlation of intermediate feature spaces, we show that the residual Squeeze and Excitation modules help the model to reduce redundancy in the latent features. Therefore, the proposed model advances the state-of-the-art in the augmentation of clinical audio data and improves the data bottleneck for the design of clinical acoustic sensing systems.



\end{abstract}

\section{Introduction}

Medical acoustic sensing systems utilize air- and structure-borne acoustic signals that can be captured in a medical environment, such as vibration signals from surgical tools captured with contact microphones \cite{seibold2021femoralstem} or sounds acquired with air-borne microphones directly from the area of operation \cite{goossens2020acoustic}, to provide guidance and support in medical interventions and diagnostics. Because acoustic signals can be captured non-invasively and radiation-free, and the systems are low-cost and easy-to-integrate, acoustic sensing has great potential for the design of multimodal sensing paradigms for the support of human surgeons, surgical diagnostics, robotic surgery, or to analyze surgical workflow. Hereby, acoustic sensing can be used to obtain measurements for applications where conventional medical computer aided support systems are limited, for example for the assessment of implant-bone press-fit which is impossible to obtain using imaging or navigation \cite{seibold2021femoralstem,goossens2020acoustic} or to complement the limitations of medical imaging for the assessment of implant loosening \cite{arami2018knee} or cartilage degeneration \cite{kim2009vag}. 

Exemplary applications for the successful application of acoustic sensing in medical interventions are error prevention in orthopedic surgery by analyzing drill vibrations to detect drill breakthrough \cite{seibold2021realtime}, 
the evaluation of implant seating during insertion of the femoral stem component in Total Hip Arthroplasty (THA) \cite{goossens2020acoustic,seibold2021femoralstem}, or the guidance of the insertion process of surgical needles using structure-borne acoustic signals acquired from the distal end of the medical device \cite{illanes2018novel}. Also in medical diagnostics, acoustic signals have been successfully employed, e.g. for cough detection \cite{alqudaihi2021cough} or the examination of heart sounds \cite{giordano2019heart}.

In the recent years, deep learning-based analysis methods have outperformed classical signal processing and machine learning techniques for the processing of acoustic signals \cite{purwins2019deep} which has also been applied in the medical domain in first use cases and showed promising performance improvements \cite{seibold2021femoralstem,seibold2021realtime}. While these methods are very powerful, they require large-scale high-quality training data to achieve superior performance and generalization to unseen cases. One of the main challenges for medical applications, however, is the limited availability of large amounts of data due to the limited access to the real surgical environment, expensive acquisition of realistic data, and clinical requirements and regulations. While in the non-medical domain of audio deep learning research, large-scale audio datasets, such as the Librispeech dataset for speech recordings \cite{panayotov2015librispeech} or the UrbanSound-8K dataset for environmental audio \cite{salamon2014urbansound8k}, are publicly available, the medical domain is lacking large-scale community data for the development of medical acoustic sensing systems. Therefore, especially in the medical domain, data augmentation is a valuable tool to artificially increase the size of a training data set to increase the diversity of training examples and stabilize the training process. To address this issue, we published a medical audio dataset in a previous work which contains acoustic signals recorded in the real operating room during THA procedures which resemble typical surgical actions such as hammering, drilling, or sawing \cite{seibold2022conditional} and proposed a data augmentation method based on a conditional generative adversarial network.

However, we note that several studies report that deep networks tend to learn redundant features due to the huge model capacity \cite{liu2021discrimination} \cite{joshi2022restoration} \cite{singh2020leveraging}. Channel attention has been successfully exploited to model channel level dependencies and facilitate learning of less redundant features \cite{hu2018se} \cite{roy20223d} \cite{choi2020channel} and subsequently improved model performance. Motivated by these observations, in this paper, we demonstrate that due to the huge number of model parameters, conditional generative adversarial network (cWGAN-GP \cite{seibold2022conditional}) learns redundant features. To combat this, we introduce a channel-wise attention mechanism in the generator sub-network through the implementation of Squeeze \& Excitation \cite{hu2018se} block and residual skip connections \cite{he2016resnet}. We provide visualizations that signify the reduced redundancy and subsequently, improved quality of generated mel spectrograms samples quantified by a custom version of the Fréchet Inception Distance \cite{heusel2017fid}. As a result, the present work advances the state-of-the-art in data augmentation for the emerging field of medical acoustic sensing and addresses the important issue of data limitations for medical deep learning-based systems.


\section{Materials and Methods}

\subsection{Data set, Preprocessing, and Benchmark Augmentations}

We use a publicly available data set\footnote{The data set can be obtained from: \href{https://rocs.balgrist.ch/open-access/}{https://rocs.balgrist.ch/open-access/}} \cite{seibold2022conditional} recorded during real Total Hip Arthroplasty surgeries and contains sounds of the typical surgical actions that are performed during the intervention and roughly resemble the different phases of the procedure. The data set includes 568 recordings with a length of \SIrange{1}{31}{\s} and the following distribution: $n_{raw, Adjustment} = 68$, $n_{raw, Coagulation} = 117$, $n_{raw, Insertion} = 76$, $n_{raw, Reaming} = 64$, $n_{raw, Sawing} = 21$, and $n_{raw, Suction} = 222$.

We compute mel spectrograms, a feature representation for audio signals that obtains state-of-the-art results for deep learning-based audio signal processing systems \cite{purwins2019deep}, using non-overlapping sliding windows which results in the following sample distribution for the entire data set: $n_{spec, Adjustment} = 494$, $n_{spec, Coagulation} = 608$, $n_{spec, Insertion} = 967$, $n_{spec, Reaming} = 469$, $n_{spec, Sawing} = 160$, and $n_{spec, Suction} = 899$.
Mel spectrograms provide a compact representation, capture time- and frequency-domain aspects about a signal and can be computed from a raw waveform by first computing the Short-time Fourier Transform (STFT) $X$ and then filtering the resulting spectra using a triangular filter bank spaced evenly on the mel scale \cite{stevens1937mel} to compute the mel spectrogram $X_{mel}$. All spectrograms computed within the present work have dimensions $64 \times 64$ and are normalized using the formula $X_{norm} = (X_{mel} - \mu) / \sigma$ where $\mu$ is the mean and $\sigma$ is the standard deviation computed over the entire data set. 

A number of data augmentation techniques for acoustic signals have been proposed in prior research, among them classical raw signal based methods like adding noise, time stretching, and pitch shifting, as well as spectrogram-based methods, e.g. SpecAugment \cite{park2019specaugment}. Furthermore, we compare the results of the proposed data augmentation framework with the results reported in our previous work \cite{seibold2022conditional} in which a standard convolutional conditional generative adversarial network with Wasserstein Loss with Gradient Penalty regularization \cite{gulrajani2017improved} was employed.

\subsection{Proposed Data Augmentation Method}

\begin{figure}
    \centering
    \includegraphics[width=\textwidth]{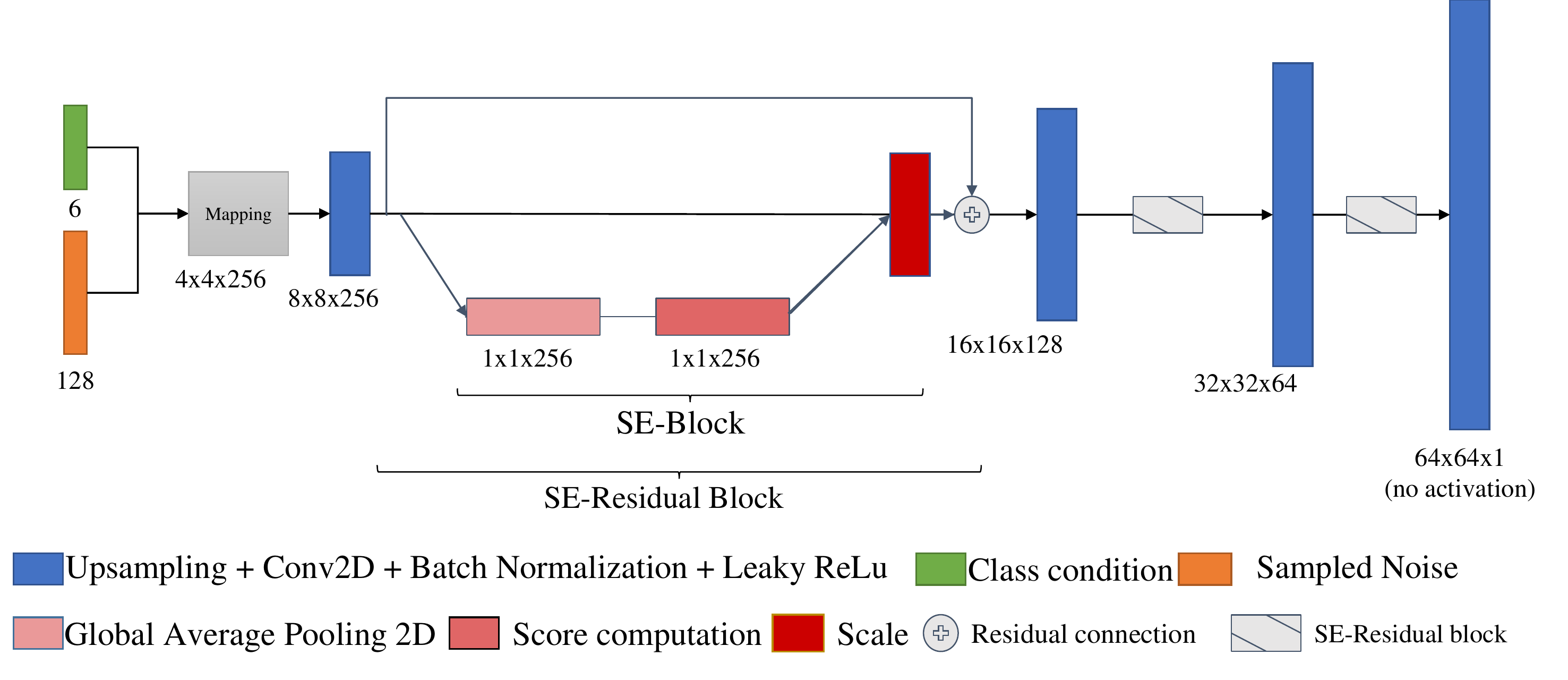}
    \caption{The schematic illustrates the structure of the proposed SE-ResNet generator for the generation of synthetic mel spectrograms.}
    \label{fig:generator}
\end{figure}

The architecture of the proposed GAN's generator is depicted in the Figure \ref{fig:generator}. It consists of 4 convolutional upsampling blocks followed by a squeeze-and-excitation block with a residual connection, a technique originally proposed in by Hu et al. \cite{hu2018se}. The Squeeze and Excitation block consists of a global average pooling layer, which allows to \textit{squeeze} global information to channel descriptors, a re-calibration part, which acts as a channel-wise attention mechanism and allows to capture channel-wise relationships in a non-mutually-exclusive way. The last operation scales the input's channels by multiplying them with the obtained coefficients. The Squeeze and Excitation mechanisms adds two fully connected layers with a ReLU activation function in between and a sigmoid function applied in the end as shown in the equation \ref{excitation}.

\begin{equation}
    s = F_{ex}(z, W) = \sigma(g(z, W)) = \sigma(W_2\delta(W_1z))
    \label{excitation}
\end{equation}

Here the variables $W_1$ and $W_2$ have the dimensions $(\frac{C}{r} \times C)$ and $(C \times \frac{C}{r})$, respectively, $\sigma$ is the sigmoid function and $\delta$ refers to a ReLU activation. The value of $r$ is a hyperparameter and for our method it was chosen equal to 16 in an empirical manner.

The generator has an overall of $1,537,316$ parameters. For the discriminator we use a fully convolutional network architecture with a total of $4,321,153$ parameters analogous to our own previous work \cite{seibold2022conditional}. Both the generator and discriminator employ the LeakyReLU non-linear activation function throughout the whole network structure. As a loss function the Wasserstein Loss with Gradient Penalty (GP) was chosen with GP weight equal to $\lambda=10$. For both the generator and the discriminator, we utilized the Adam optimizer with a learning rate of $\lambda = 5\times10^{-4}$. 
The discriminator was trained for 5 extra steps per epoch. The implementation and training of all reported results were done using Tensorflow/Keras 2.6 using a Google Cloud instance running a single NVIDIA T4 GPU.

The determination of when to stop the training process is notoriously difficult for the training of GANs. To assess the quality of the generated samples, we repeatedly compute a custom version of the Fréchet Inception Distance \cite{heusel2017fid} which is computed based on the features of the last convolutional layer of a ResNet-18 \cite{he2016resnet} pre-trained on the THA data set published in \cite{seibold2022conditional}. The training process is stopped when the lowest FID is observed which is computed using the equation \ref{fid}, where $\mu_r$ and $\mu_g$ is the feature-wise mean of the real and generated spectrograms, $C_r$ and $C_g$ are the covariance matrices.

\begin{equation}
    FID = \|\mu_r - \mu_g\|^2 + \Tr(C_r + C_g - 2*\sqrt{C_r * C_g})
    \label{fid}
\end{equation}

\subsection{Classifier for Evaluation}

For the evaluation of the proposed improved data augmentation method we employed a ResNet-18 classifier as previously reported in \cite{seibold2022conditional} which is a standard convolutional neural network architecture for spectrogram-based audio classification tasks and has been shown to achieve state-of-the-art results in medical acoustic sensing applications \cite{seibold2021femoralstem,seibold2022conditional}. To be able to compare the results presented within this work with the previous results, we augment 
100\% synthetic samples for each class present in the data set. The classifier was trained for 20 epochs using 5-fold cross-validation technique. We used categorical cross-entropy loss with the Adam optimizer and the following hyperparameters: learning rate = $10^{-5}$, $\beta_1 = 0.9 $, $\beta_2 = 0.99$

\section{Results}
\begin{figure}
    \centering
    \includegraphics[width=\textwidth]{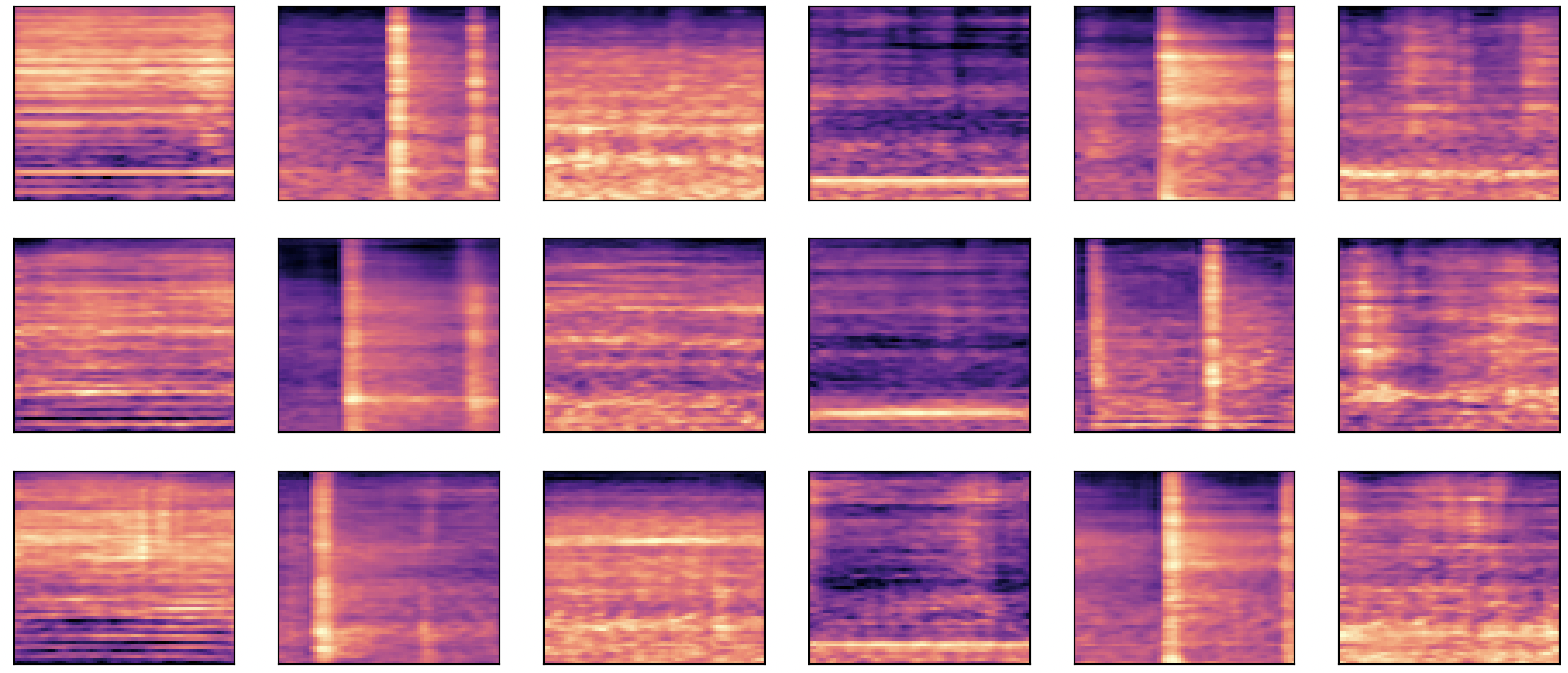}
    \caption{Log-mel spectrogram of random samples for each class (top row); log-mel spectrogram of random generated images of our proposed model (second row); log-mel spectrogram of the model proposed in the previous work\cite{seibold2022conditional} (bottom row). Respective classes from left to right: 
Sawing, Adjustment, Reaming, Coagulation, Insertion, Suction}
    \label{fig:specs}
\end{figure}

In order to visually compare the quality of the proposed model, we present per-class randomly selected ground truth data, generated spectrograms from the proposed model, and synthetic spectrograms generated from the previous augmentation framework \cite{seibold2022conditional} in figure \ref{fig:specs}.

We stopped the training by frequently monitoring the quality of the generated samples through the computation of the FID as described in equation \ref{fid} and selected the best model with the lowest FID score which was subsequently used to augment the data set by doubling the number of samples for each class, the best performing augmentation strategy identified in previous work. We report the mean Macro F1 score over a five-fold cross validation experiment in the format mean $\pm$ standard deviation. A comparison between the classifier performance with no augmentations, using classical signal- and spectrogram-processing-based methods, the method proposed in our own previous work \cite{seibold2022conditional}, and the proposed model is shown in the Table \ref{tab:resulttable}.

 \begin{table}[h!]
     \centering
     \begin{tabular}{|p{0.22\textwidth}|p{0.07\textwidth}|p{0.22\textwidth}|p{0.18\textwidth}|}
     \hline
     
          \textbf{Augmentation method} & \textbf{FID} &  \textbf{Macro\,\,F1-Score  \quad \quad \quad  (mean $\pm$ std)} & \textbf{Relative \quad improvement}  \\ 
          \hline
          \hline
    
          No augmentation & & $93.9 \pm 2.5\%$ & \\ 
          White noise & & $92.87 \pm 0.99\%$  & $-1.03\%$ \\ 
          Pitch Shift & & $94.73\pm 1.28 \%$ & $0.83 \%$ \\ 
          Time Stretch & & $95.0 \pm 1.49 \%$ & $1.1\%$ \\ 
          SpecAugment\cite{park2019specaugment}  & & $94.23  \pm 1.14 \%$ & $0.33 \%$ \\ 
          cWGAN-GP\cite{seibold2022conditional} & $3.30$ & $95.60 \pm 2.6 \%$ & $1.7\%$ \\ 
          \textbf{Our method} & $\mathbf{3.01}$ & $\mathbf{96.74 \pm 1.03 \%}$ & $\mathbf{2.84 \%}$ \\ 
    \hline
     \end{tabular}
     \caption{Comparison of different augmentation methods for clinical audio data. All reported results were obtained by applying the respective augmentation method to double the number of samples for each class of the public THA sounds data set.}
     \label{tab:resulttable}
     
 \end{table}
 
\begin{figure}
    \centering
    \includegraphics[width=\textwidth]{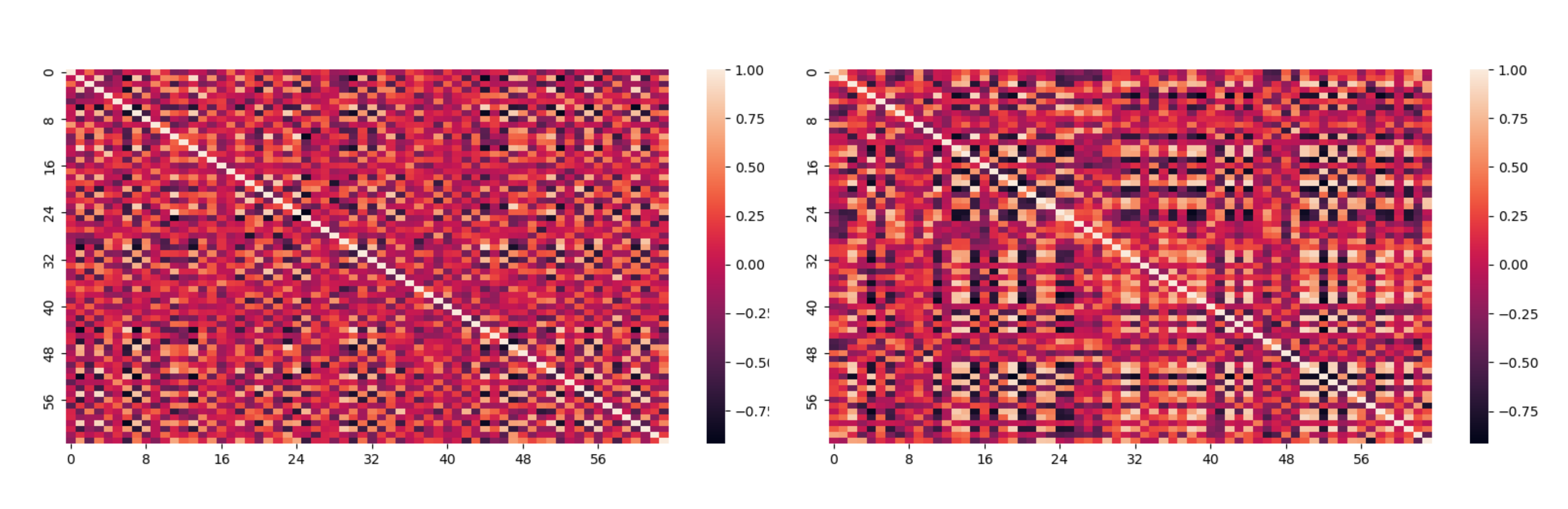}
    \caption{Sample correlation matrices of features learned by the proposed model (left column) and cWGAN-GP published in previous work \cite{seibold2022conditional}. The correlation matrices are computed from different intermediate layers of the generator network. The plots represent the correlation in the feature space after the second-last convolutional layer with dimensions 32x32x64. The significantly lower correlation values obtained after introducing Squeeze \& Excitation block demonstrate the reduced correlation among features and therefore reduced feature redundancy.}
    \label{fig:correlation}
\end{figure}

To analyze the redundancy in learned feature space of the proposed model and compare it with the previously published method, we plot the correlation matrices computed from intermediate layers of the network to analyze the redundancy of features in figure \ref{fig:correlation}. The results show that the redundancy of features is significantly reduced by introducing residual Squeeze and Excitation modules in the generator network. 

\section{Discussion}
Deep learning-based acoustic sensing has been shown to have high potential for clinical applications in diagnostics and interventional guidance, can be used for multimodal sensing to complement established assistance systems, and provide data beyond the limits of computer aided diagnostic and interventional support systems. However, to achieve state-of-the-art results, learning-based systems rely on big training data sets to generalize well for unseen cases. Obtaining these large amounts of clinical data is a common problem for the design of deep learning-based support and guidance systems in medicine. Advanced augmentation methods have been designed for medical imaging applications \cite{tirindelli2021ultrasound,shin2018augmentation} and a first method for the augmentation of clinical audio data sets has been proposed by the authors in previous work \cite{seibold2022conditional}. 

In the present work, the results show that the proposed method outperforms all previously suggested augmentation methods. In comparison to the first generative modeling based method for clinical audio data, we outperform the model by a margin of $1.14\%$ in Macro F1-Score. While this is an incremental improvement, we could significantly improve the results by only adding a total number of 11232 additional parameters which corresponds to a parameter growth of only $0.74\%$ for the generator model. Furthermore, the correlation analysis of intermediate latent features revealed that the introduced residual Squeeze and Excitation modules reduce the redundancy in the learned features of the generator model. Therefore, the proposed architecture is a highly valuable extension in the generator architecture for an improved synthetic generation of mel spectrograms. An improvement of 0.3 in the reported FID score underlines the capabilities of the proposed architectural modifications.

The proposed approach can generate any arbitrary number of samples for the classes present in the learned data set distribution and could therefore be employed to address data imbalance issues. However, in the current work we focused on improving the quality of the generated samples. Therefore, a more thorough investigation regarding the influence of different augmentation schemes using conditional generative data augmentation should be subject to future research.

By introducing a generative deep-learning method, the processing time for generating the augmentations increases in comparison to simple signal processing-based approaches. To investigate the capabilities of the proposed method, the model should be trained and evaluated on multiple relevant clinical audio data sets in future research.

\section{Conclusion}

In this work, we propose an enhanced generator architecture for conditional generative learning-based data augmentation of clinical audio data. We outperform all previously published methods and provide an in-depth analysis of the proposed modifications, residual Squeeze and Excitation modules in the generator structure. The method is able to increase the quality of synthetically generated samples by $0.3$ in terms of FID score and improves the performance of a classifier trained on the augmented data set by a margin of $2.84\%$ in terms of Macro F1-Score. All presented results are evaluated on a public data set containing sounds of a Total Hip Arthroplasty procedure which was recorded in the real operating room and evaluated using a 5-fold cross validation scheme. The obtained results show that the proposed method has great potential to improve the problem of data limitations for the design of clinical acoustic sensing systems.

\section*{Acknowledgement}
This work is part of the SURGENT project under the umbrella of Hochschulmedizin Zürich.

\bibliographystyle{elsarticle-num} 
\bibliography{bibliography}

\end{document}